%% file: root.tex
\documentclass[letterpaper, 10 pt, conference]{IEEEtran}

\IEEEoverridecommandlockouts

\usepackage[table,usenames,dvipsnames]{xcolor}
\usepackage{cite}
\usepackage[dvipsnames]{xcolor}
\usepackage{hyperref}

\usepackage{amsmath,amssymb,amsfonts, amsthm, dsfont} 
\usepackage{mathtools, bm}
\usepackage{siunitx}
\usepackage[linesnumberedhidden, lined, boxed, ruled]{algorithm2e}

\usepackage{float}
\usepackage{graphicx}
\ifCLASSOPTIONcompsoc
\usepackage[caption=false,font=normalsize,labelfont=sf,textfont=sf]{subfig}
\else
\usepackage[caption=false,font=footnotesize]{subfig}
\fi
\usepackage{tabularx, threeparttable, booktabs}
\usepackage{multirow,multicol}

\newcommand{\prl}[1]{\mathopen{}\left(#1\right)\mathclose{}}
\newcommand{\brl}[1]{\mathopen{}\left[#1\right]\mathclose{}}
\newcommand{\crl}[1]{\mathopen{}\left\{#1\right\}\mathclose{}}

\title{\vspace{6mm}\LARGE \bf
Feedback Enhanced Motion Planning for Autonomous Vehicles
}

\author{
  Ke~Sun, Brent~Schlotfeldt, Stephen~Chaves, Paul~Martin, Gulshan~Mandhyan, and~Vijay~Kumar
  \thanks{ Ke Sun, Brent Schlotfeldt, and Vijay Kumar are with GRASP Lab, University of Pennsylvania, Philadelphia, PA 19104, USA, {\tt\small\{sunke, brentsc, kumar\}@seas.upenn.edu}. Stephen Chaves, Paul Martin, and Gulshan Mandhyan are with Qualcomm Technologies Inc., Philadelphia, PA 19146, USA, {\tt\small\{schaves, pdmartin, gmandhya\}@qti.qualcomm.com}. We gratefully acknowledge the support of Qualcomm Research who sponsored this work.}
}

\begin{document}

\maketitle
\thispagestyle{empty}
\pagestyle{empty}

\begin{abstract}
  In this work, we address the motion planning problem for autonomous vehicles through a new lattice planning approach, called Feedback Enhanced Lattice Planner (FELP). Existing lattice planners have two major limitations, namely the high dimensionality of the lattice and the lack of modeling of agent vehicle behaviors. We propose to apply the Intelligent Driver Model (IDM)~\cite{treiber2013traffic} as a speed feedback policy to address both of these limitations. IDM both enables the responsive behavior of the agents, and uniquely determines the acceleration and speed profile of the ego vehicle on a given path. Therefore, only a spatial lattice is needed, while discretization of higher order dimensions is no longer required. Additionally, we propose a directed-graph map representation to support the implementation and execution of lattice planners. The map can reflect local geometric structure, embed the traffic rules adhering to the road, and is efficient to construct and update. We show that FELP is more efficient compared to other existing lattice planners through runtime complexity analysis, and we propose two variants of FELP to further reduce the complexity to polynomial time. We demonstrate the improvement by comparing FELP with an existing spatiotemporal lattice planner using simulations of a merging scenario and continuous highway traffic. We also study the performance of FELP under different traffic densities.
\end{abstract}

\input{tex/introduction}

\input{tex/problem_formulation}
\input{tex/algorithm}

\input{tex/experiments}

\input{tex/conclusion}




\bibliographystyle{IEEEtran}
\bibliography{ref}

\end{document}

%% file: tex/introduction.tex
\section{Introduction}
\label{sec: introduction}
Motion planning for self-driving vehicles remains challenging, not only because dynamically feasible trajectories for the ego vehicle should be generated in a structured environment, but also because the behavior of agent vehicles has to be predicted to avoid collisions.
A common approach to address the motion planning problem is based on decomposing it into two sub-problems, namely behavior planning and trajectory planning~\cite{paden2016survey}. The underlying heuristics is that solving the two sub-problems is easier than solving the original motion planning problem directly. Behavior planners~\cite{hubmann2017decision, galceran2017multipolicy, sunberg2017value, hubmann2018belief} generally serve two purposes: 1) generating high-level commands, such as lane keeping or changing; 2) predicting the intentions of the agent vehicles. Trajectory planners~\cite{ziegler2014trajectory, fan2018baidu, chen2019autonomous} then solve a smooth trajectory often through convex optimization, utilizing the output from the behavior planner.

While decomposition approaches achieve promising results, two issues remain hard to resolve in this framework. First, as pointed out by McNaughton~\cite{mcnaughton2011motion}, behavior planning often relies on a flawed model of the underlying trajectory planning. This mismatch may lead to unstable or, worse, infeasible trajectory optimization. Sadat~\cite{sadat2019jointly} discusses the second limitation of the decomposition framework: behavior and trajectory planners optimize different objective functions. As a result, changes in the objective function of behavior planning may have a negative impact on the final trajectory, which then requires re-tuning or re-designing the objective function for the trajectory planner.
Lattice planners~\cite{claussmann2019review} could be a potential solution to resolve the issues. Instead of decomposing the motion planning problem into behavior and trajectory planning, lattice planners address the problem directly by discretizing the spatiotemporal space and finding the optimal trajectory through graph search.

State lattice approaches~\cite{pivtoraiko2009differentially} are originally designed for rover-like vehicles operating in unstructured environments. Ziegler~\cite{ziegler2009spatiotemporal} extends the concept to a spatiotemporal lattice by introducing time as the extra dimension. Although it is efficient to search for an optimal solution within the spatiotemporal lattice, constructing the lattice is time-consuming. This limitation makes the method impractical for continuous autonomous driving tasks, where repetitive lattice construction is required as the local environment around the ego vehicle changes.

McNaughton~\cite{mcnaughton2011motion} leverages the fact that the path and velocity of a trajectory can be decoupled~\cite{kant1986toward}. In~\cite{mcnaughton2011motion}, position is discretized explicitly. Clothoid paths~\cite{kelly2003reactive} are used to connect positions close to each other. For each path, trajectories are obtained by applying different constant accelerations. Hence, instead of explicit discretization, velocity and time are induced by the starting state and the applied acceleration. Generating velocity profiles with constant acceleration echos another regime of creating a state lattice~\cite{rufli2010design}, where the state lattice is no longer obtained by explicitly discretizing the state space, but rather induced by motion primitives obtained by discretizing the control space. The performance of the algorithm in~\cite{mcnaughton2011motion} depends heavily on the set of constant accelerations. An overly coarse set results in jerky trajectories, while a fine set significantly increases running time.

Recently, Ajanovic~\cite{ajanovic2018search} proposes another way of constructing the spatiotemporal lattice in the same spirit as~\cite{mcnaughton2011motion}. In addition to the differences in the constraints and types of paths, the velocity dimension is discretized explicitly in~\cite{ajanovic2018search}, leaving position and time to be induced from the initial state and the applied constant acceleration. Since the algorithm structure remains the same, the method in~\cite{ajanovic2018search} has the same computation complexity as~\cite{mcnaughton2011motion} and has the same limitation caused by a finite set of constant accelerations.

Limitations of lattice planners, such as those in~\cite{ziegler2009spatiotemporal, mcnaughton2011motion, ajanovic2018search}, are notable. As discussed above, lattice planners require discretization, either explicitly or implicitly, of the spatiotemporal space which is often of high dimensionality. The high dimensionality makes the lattice expensive to be repeatedly constructed to adapt to changing environments. Another limitation of existing lattice planners is the lack of modeling of agent vehicles' responsive behavior. In~\cite{ziegler2009spatiotemporal, mcnaughton2011motion, ajanovic2018search}, agent vehicles are assumed to maintain constant velocity. Compared with decomposition frameworks, behavior planning provides such modeling, where advanced prediction of agent vehicles' behavior~\cite{hubmann2017decision, galceran2017multipolicy, sunberg2017value} is possible at the cost of abstracting the controls to only a few high-level maneuvers. However, it is not clear how to apply these methods to lattice approaches.

Additionally, the efficiency of the planning algorithms relies on the representation of the local environment, \textit{i.e.} the map. The map should satisfy three requirements: 1) reflect the geometric structure of the local environment; 2) encode the traffic rules adhering to the road (e.g. an exit-only lane); and 3) be constructed, accessed and updated efficiently. Widely used map formats include OpenStreetMap~\cite{haklay2008openstreetmap}, OpenDrive\cite{opendrive}, Lanelet~\cite{bender2014lanelets} and Lanelet2~\cite{poggenhans2018lanelet2}, which are designed for large scale environments. Although these can be used for motion planning in theory, repeatedly retrieving information from such map representations can be time-consuming. Few papers in the motion planning literature discuss the map representation. \cite{mcnaughton2011motion} briefly reports that the local environment is represented with an occupancy grid map. However, occupancy grid maps~\cite[Ch.9]{thrun2000probabilistic} are not designed for structured environments, satisfying none of the above requirements.

\textbf{Contributions:} In this work, we propose a Feedback Enhance Lattice Planner (FELP), addressing the motion planning problem of self-driving vehicles in highway scenarios.
Our major contributions are summarized as follows:

First, we use an Intelligent Driver Model (IDM)~\cite[Ch.11]{treiber2013traffic} as the feedback policy to control the speed for both the ego and agent vehicles. Given a speed feedback policy for the ego, the velocity at the end of a trajectory and the time used to execute the trajectory are determined uniquely knowing the path and the starting state. Thus, discretization of acceleration, velocity, and time dimension is no longer required. The lattice remains in the 2-D spatial space. Meanwhile, IDM provides an efficient way to model the responsive behaviors for agent vehicles, which integrates seamlessly with lattice planning approaches.

Second, we propose a directed-graph map representation. The proposed representation satisfies all of the previously discussed map requirements. In addition to representing the static environment, vehicles can also be registered onto the map, simplifying the process of collision checking and identifying the relative positions of vehicles.

Finally, we show that the runtime complexity of FELP is significantly reduced compared to the lattice planners in~\cite{ mcnaughton2011motion, ajanovic2018search}. While the complexity still grows exponentially with the spatial planning horizon, we propose two variants of FELP that bring down the complexity to polynomial time.

%% file: tex/problem_formulation.tex
\section{Problem Formulation}
\label{sec: problem formulation}
We start the problem formulation by considering the dynamics of a single vehicle. As in~\cite{kelly2003reactive}, a vehicle is modeled as a unicycle,
\begin{equation}
  \label{eq: dynamics of a vehicle}
  \begin{gathered}
    \dot{\bm{x}}(t)  =
    \prl{\dot{x}(t), \dot{y}(t), \dot{\theta}(t), \dot{v}(t)}^\top =
    f(\bm{x}(t), \bm{u}(t)) = \\
    \prl{v(t) \cos\theta(t), v(t) \sin\theta(t), v(t) \kappa(t), a(t)}^\top.
  \end{gathered}
\end{equation}
The state, $\bm{x}=\prl{x, y, \theta, v}^\top \in \mathcal{X}$, consists of 2-D position $(x, y)$, orientation $\theta$, and speed $v$. The control $\bm{u}=\prl{\kappa, a}^\top \in \mathcal{U}$ is composed of curvature $\kappa$ and acceleration $a$. More precisely, $a$ is the norm of tangential acceleration.

By stacking the dynamics of a single vehicle in Eq.~\eqref{eq: dynamics of a vehicle}, the dynamics of the local traffic can be constructed as,
\begin{equation}
  \label{eq: dynamics of traffic}
  \dot{\bm{x}}_s(t) =
  \begin{pmatrix}
    \dot{\bm{x}}_e(t) \\ \dot{\bm{x}}_0(t) \\ \vdots \\ \dot{\bm{x}}_{M-1}(t)
  \end{pmatrix} =
  \begin{pmatrix}
    f\prl{\bm{x}_e(t), \bm{u}_e(t)} \\
    f\prl{\bm{x}_0(t), \bm{u}_0(t)} \\
    \vdots                          \\
    f\prl{\bm{x}_{M-1}(t), \bm{u}_{M-1}(t)}
  \end{pmatrix},
\end{equation}
where $\bm{x}_e$ is the ego state, $\bm{x}_i, i=0, 1, \dots, M-1$ are the states of agent vehicles. Together, $\bm{x}_s=\prl{\bm{x}_e^\top, \bm{x}_0^\top, \dots, \bm{x}_{M-1}^\top}^\top \in \mathcal{X}^{M+1}$ is the traffic state.

With the assumption that agent vehicles are lane followers, curvature for the agent vehicle $i$ can be generated based on the lane geometry, denoted as $\kappa\prl{\bm{x}_i}:\mathcal{X}\mapsto\mathbb{R}$. By modulating vehicle speed with IDM, acceleration for the agent vehicle $i$ can be generated by a feedback function $\xi\prl{\bm{x}_i, \bm{y}_i, \bm{\lambda}_i}:\mathcal{X}\times\mathcal{X}\times \Lambda\mapsto\mathbb{R}$. In $\xi\prl{\cdot}$, $\bm{y}_i$ is the state of the leading vehicle of agent vehicle $i$. $\bm{\lambda}_i\in\Lambda$ consists of the IDM hyper-parameters for agent vehicle $i$. In this work, we assume that $\bm{\lambda}_i$ are known for all agents. Therefore, $\xi\prl{\bm{x}_i, \bm{y}_i, \bm{\lambda}_i}$ can be simplified to $\xi\prl{\bm{x}_i,\bm{y}_i}:\mathcal{X}\times\mathcal{X}\mapsto\mathbb{R}$, embedding the IDM hyper-parameters as constants in $\xi\prl{\cdot}$. Combining $\kappa\prl{\cdot}$ and $\xi\prl{\cdot}$ gives the feedback policies for agent vehicles,
\begin{equation}
  \label{eq: agent feedback policy}
  \pi_i\prl{\bm{x}_s} = \prl{\kappa\prl{\bm{x}_i}\; \xi\prl{\bm{x}_i, \bm{y}_i}}^\top.
\end{equation}
The traffic dynamics in Eq.~\eqref{eq: dynamics of traffic} is updated as a result of the introduced feedback policies in Eq.~\eqref{eq: agent feedback policy},
\begin{equation}
  \label{eq: dynamics of traffic with agent feedback policy}
  \dot{\bm{x}}_s(t) =
  \begin{pmatrix}
    \dot{\bm{x}}_e(t) \\ \dot{\bm{x}}_0(t) \\ \vdots \\ \dot{\bm{x}}_{M-1}(t)
  \end{pmatrix} =
  \begin{pmatrix}
    f\prl{\bm{x}_e(t), \bm{u}_e(t)} \\
    f\prl{\bm{x}_0(t), \pi_0(\bm{x}_s)} \\
    \vdots                          \\
    f\prl{\bm{x}_{M-1}(t), \pi_{M-1}(\bm{x}_s)}
  \end{pmatrix},
\end{equation}
which is of the form $\dot{\bm{x}}_s(t) = f_s(\bm{x}_s(t), \bm{u}_e(t))$. In practice, the behavior of agents may not exactly follow the assumed model in Eq.~\eqref{eq: agent feedback policy}. In this work, FELP is applied in the framework of Receding Horizon Control (RHC), which re-plans at a fixed frequency to utilize the latest traffic state.

With the dynamics of the traffic system, the motion planning problem for autonomous driving can be formulated as the following optimal control problem,
\begin{equation}
  \label{eq: general autonomous driving problem}
  \begin{aligned}
    \min_{\bm{u}_e(t),\; t_f}& \int_{t_0}^{t_f} c_g\prl{\bm{x}_s(\tau), \bm{u}_e(\tau)} d\tau + c_t\prl{\bm{x}_s(t_f)} \\
    \text{subject to} & \quad
    \begin{aligned}[t]
      &\text{(1) }\dot{\bm{x}}_s(t) = f_s(\bm{x}_s(t), \bm{u}_e(t)) \\
      &\text{(2) collision avoidance} \\
      &\text{(3) traffic rules}.
    \end{aligned}
  \end{aligned}
\end{equation}
In Eq.~\eqref{eq: general autonomous driving problem}, $t_f$ is the free final time, $c_g:\mathcal{X}^{M+1}\times\mathcal{U}\mapsto\mathbb{R}$ is the running cost, and $c_t:\mathcal{X}^{M+1}\mapsto\mathbb{R}$ is the terminal cost. Note that IDM ensures collision-free driving assuming no acceleration constraints. However, in our implementation, the vehicle acceleration is bounded by physical limits. Therefore, an additional collision avoidance constraint is still required.


\input{./tex/problem_formulation/ego_motion_primitives}
\input{./tex/problem_formulation/map_representation}

\subsection{Consolidated Problem Formulation}
\label{subsec: consolidated problem formulation}
\begin{figure}[t]
  \centering
  \subfloat[]{\includegraphics[width=\columnwidth]{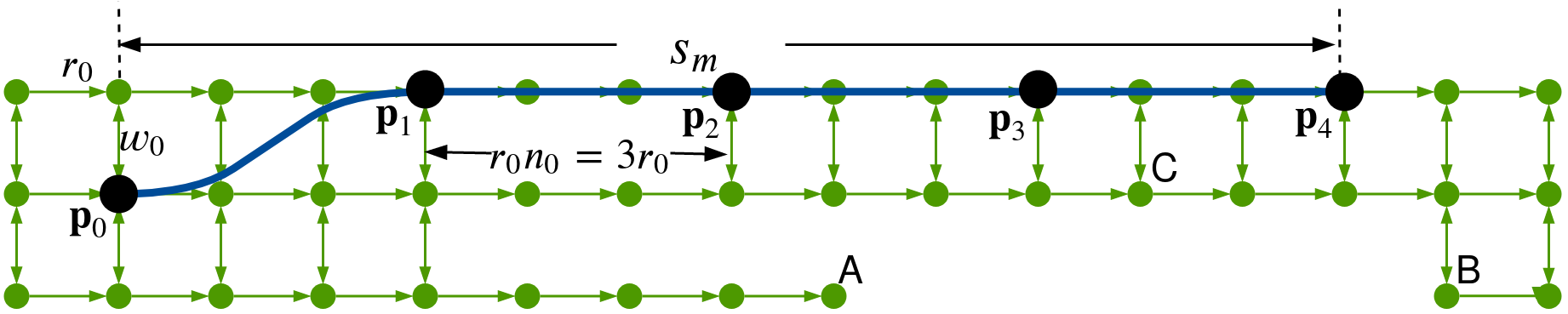}\label{subfig: a feasible solution}} \\
  \subfloat[]{\includegraphics[width=\columnwidth]{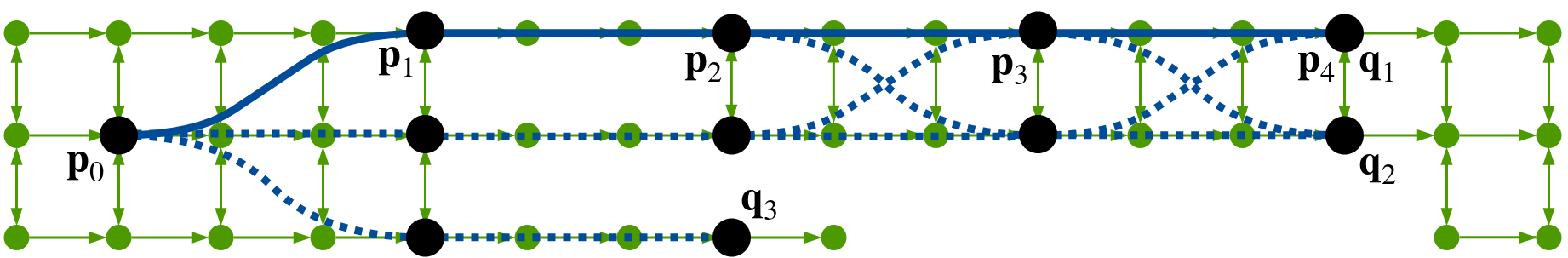}\label{subfig: the optimal solution}}
  \caption{(a) shows a feasible solution (\textcolor{blue}{blue} paths) to the problem in Eq.~\eqref{eq: unconstrained autonomous driving problem} using the directed-graph map in \figurename~\ref{fig: directed graph map}. As in \figurename~\ref{subfig: illustration for functions d phi}, $w_0$ and $r_0$ are the metric length of the lateral and longitudinal edges of the map. In this example, the planning horizon $s_m=12r_0$. Longitudinal arclength of paths is $3r_0$, \textit{i.e.} $n_0$ in Eq.~\eqref{eq: unconstrained autonomous driving problem} is $3$. (b) shows the paths (\textcolor{blue}{blue}) constructed in the search process of Alg.~\ref{alg: graph search for optimal motion primitives}. $\bm{q}_0$, $\bm{q}_1$, and $\bm{q}_2$ are terminal waypoints. The traffic states at these waypoints constitute the terminal set, $\mathcal{T}$, in Alg.~\ref{alg: graph search for optimal motion primitives}. The solid path sequence, with end points marked from $\bm{p}_0$ to $\bm{p}_4$, is the possible optimal solution.}
  \label{fig: motion primitive solutions}
\end{figure}

With definitions for the ego motion primitives and the directed-graph map, we are able to consolidate the problem formulation in Eq.~\eqref{eq: general autonomous driving problem}. First, we reformulate Eq.~\eqref{eq: general autonomous driving problem} without the constraints of collision avoidance and traffic rules,
\begin{equation}
  \label{eq: unconstrained autonomous driving problem}
  \begin{aligned}
    \min_{\bm{p}_k,\, k=1, 2, \dots}& \sum_{k} c_s\prl{\bm{p}_k} + c_t\prl{\bm{x}_s(t_f)} \\
    \text{subject to} & \quad
    \begin{aligned}[t]
      &\bm{p}_0 = \prl{x_e(t_0), y_e(t_0), \theta_e(t_0), \kappa_e(t_0)}^\top\\
      &\bm{p}_k \in \mathcal{V},\; s(\bm{p}_k) \leq s_m \\
      &s(\bm{p}_k) - s(\bm{p}_{k-1}) = r_0\cdot n_0\\
      &|l(\bm{p}_k) - l(\bm{p}_{k-1})| \leq w_0 \\
    \end{aligned}
  \end{aligned}
\end{equation}
Instead of directly optimizing $\bm{u}_e(t)$, the optimization is over the motion primitives of the ego, represented as the end points, $\bm{p}_k$'s, of paths. Because optimization variables have changed, the objective function must be updated. Integrating the running cost $c_g$ gives the stage cost $c_s$,
\begin{equation*}
  c_s(\bm{p}) = \int_{t_1}^{t_2} c_g(\bm{x}_s(\tau), \bm{\pi}_{e, \bm{p}}(\bm{x}_s(\tau))) d\tau,
\end{equation*}
where $t_1$ and $t_2$ are the start and end time for executing the motion primitive represented by $\bm{p}$. The final time $t_f$ is no longer a variable to be explicitly optimized. Instead, $t_f$ is implicitly determined by the traffic dynamics and the selected sequence of motion primitives. The first constraint in Eq.~\eqref{eq: unconstrained autonomous driving problem} fixes $\bm{p}_0$ at the initial state of the ego, which is not necessarily at a vertex in the directed-graph map. $\bm{p}_k \in \mathcal{V}$ ensures the end points of paths are vertices in the map. $s(\bm{p}_k) \leq s_m$ sets the spatial planning horizon to $s_m$. The remaining two constraints instruct the selection of motion primitives. Paths of motion primitives are of constant arclength $r_0\cdot n_0$ in the longitudinal direction of the road. The constant arclength of paths is defined using $r_0$, the longitudinal resolution of the map, and $n_0$, the length of path in terms of edges in the longitudinal direction. In addition, each motion primitive should perform at most one lane change. We note that the last two constraints in Eq.~\eqref{eq: unconstrained autonomous driving problem} only provide one possible way of selecting motion primitives. Depending on the computation budget, different methods in selecting motion primitives can be applied. The same problem formulation should fit seamlessly. \figurename~\ref{subfig: a feasible solution} shows a feasible solution to the problem in Eq.~\eqref{eq: unconstrained autonomous driving problem}.

Thanks to the functions defined on the directed-graph map, the constraints of collision avoidance and traffic rules adhering to the road structure can be formulated. Collision avoidance means vehicles should not overlap,
\begin{equation}
  \label{eq: collision avoidance constraint}
  \begin{gathered}
    \mathcal{O}(\bm{x}_e(t)) \cap \mathcal{O}(\bm{x}_i(t)) = \emptyset \\
    t\in\brl{t_0, t_f},\quad i \in \crl{0, 1, \dots, M-1}.
  \end{gathered}
\end{equation}
In Eq.~\eqref{eq: collision avoidance constraint}, collision checking among agent vehicles is ignored. Speeds of agent vehicles are modulated with IDM. As long as initial states are reasonable, collision among agent vehicles rarely happens as we experience in the experiments. Additionally, the constraint of traffic rules adhering to the road are,
\begin{gather}
  \label{eq: traffic rule constraint 1}
  \phi(\bm{p}_{k-1}, \bm{p}_k) = \frac{|l(\bm{p}_k) - l(\bm{p}_{k-1})|}{w_0} \cdot n_0 + 1\\
  \label{eq: traffic rule constraint 2}
  d(\bm{p}_{k-1}, \bm{p}_k) \leq r_0\cdot n_0 + w_0.
\end{gather}
Eq.~\eqref{eq: traffic rule constraint 1} prevents illegal lane changes (\textit{e.g.} $C$ to $B$ in \figurename~\ref{subfig: a feasible solution}). Eq.~\eqref{eq: traffic rule constraint 2} forbids paths connecting waypoints on the same lane of a discontinuous road (\textit{e.g.} $A$ to $B$ in \figurename~\ref{subfig: a feasible solution}).

To complete the problem formulation, we comment on the cost functions implemented in our work. Running cost $c_g$ includes the acceleration and headway of the ego reflecting both comfort and safety. Since IDM is used to model agent vehicles, it is implicitly assumed that agents would always yield to the ego. To avoid inconsiderate behavior, braking of agents are also included in the running cost. Therefore, aggressive maneuvers of the ego are discouraged. Terminal cost depends on the difference between the terminal and desired speed and travelled distance. The cost of the travelled distance helps the ego avoid exit-only lanes. For example, in \figurename~\ref{subfig: the optimal solution}, the path option ending at $\bm{q}_3$ is likely to be avoided because of the short travelling distance.

%% file: tex/problem_formulation/ego_motion_primitives.tex
\subsection{Motion Primitives of the Ego Vehicle}
\label{subsec: motion primitives of the ego vehicle}
In this work, the set of motion primitives are feedback control policies instead of open-loop policies as in~\cite{rufli2010design}. Similar to Eq.~\eqref{eq: agent feedback policy}, the $j^{th}$ motion primitive for the ego is of the form,
\begin{equation}
  \label{eq: ego feedback policy}
  \pi_{e, j}(\bm{x}_s) = \prl{\kappa_{e, j}\prl{\bm{x}_e}\; \xi\prl{\bm{x}_e, \bm{y}_e}}.
\end{equation}
Some differences between Eq.~\eqref{eq: ego feedback policy} and Eq.~\eqref{eq: agent feedback policy} should be noted. First, $\kappa_{e, j}\prl{\cdot}$ is not determined by the lane geometry, but rather a specific path to be followed by the ego. Second, as the ego may change lanes, $\bm{y}_e$, the ego's leading vehicle, should be determined with special care. In this work, we define the leading vehicle of the ego as the agent which is on the same lane as the front of the ego. For example, if the ego is changing lanes, it switches its leader once its front bumper passes the lane boundary.

A few observations can help simplify the representation of the motion primitives in Eq.~\eqref{eq: ego feedback policy}. First, function $\xi\prl{\cdot}$ is the same across all motion primitives. This can be potentially removed from the representation of a motion primitive and embedded in the traffic dynamics in Eq.~\eqref{eq: dynamics of traffic with agent feedback policy}. Second, the path for each motion primitive, determining $\kappa_{e, j}\prl{\cdot}$, can be constructed knowing the bounding conditions (position, orientation, and curvature at the two end points). In this work, we use the optimization method from~\cite{kelly2003reactive} to construct such paths. Considering an ego-centric frame where the starting point of a path is fixed at the origin, the path can be determined by only providing the end point $\bm{p}=\prl{x, y, \theta, \kappa}^\top\in\mathbb{R}^4$. Combining the two observations, the set of motion primitives $\mathcal{M}$ is simply a collection of end points of paths, \textit{i.e.} $\mathcal{M}=\crl{\bm{p}_j\in\mathbb{R}^4, j=0, 1, \dots}$ with each $\bm{p}_j$ representing $\bm{\pi}_{e, j}$ in Eq.~\eqref{eq: ego feedback policy}.

Given the representation of $\mathcal{M}$, only a spatial lattice, instead of spatiotemporal lattice, is required. More specifically, discretization over acceleration, velocity, or time is no longer needed. Acceleration and velocity of the ego and the time to complete a given path are determined by the starting traffic state and the ego speed feedback policy $\xi\prl{\cdot}$. Note that although $\bm{p}_j\in\mathcal{M}$ is in $\mathbb{R}^4$, only the longitudinal and lateral dimension of the road needs to be discretized. As in~\cite{mcnaughton2011motion}, orientation and curvature at $\bm{p}_j$ are determined by the road geometry in order to ensure the path is conformal to the road.

%% file: tex/problem_formulation/map_representation.tex
\subsection{Directed-graph Map Representation}
\label{subsec: directed graph map representation}

\begin{algorithm}[t]
  \DontPrintSemicolon
  \SetCommentSty{emph}
  \KwIn{
    $\bm{p}_0$: the starting waypoint.\\
    \hspace{12mm}$r_m$: longitudinal range of the map.\\
    \hspace{12mm}$r_0$: longitudinal resolution of the map.}
  \KwOut{$\mathcal{G}=(\mathcal{V}, \mathcal{E})$}

  \SetKwFunction{frontWaypoints}{frontWaypoints}
  \SetKwFunction{leftWaypoint}{leftWaypoint}
  \SetKwFunction{rightWaypoint}{rightWaypoint}
  \SetKwFunction{onRoute}{onRoute}
  \SetKwFunction{pop}{pop}
  \SetKwFunction{range}{range}

  $\text{\range{$\bm{p}$}} \leftarrow 0$\;
  $\mathcal{E} \leftarrow \emptyset$,
  $\mathcal{V} \leftarrow \{\bm{p}_0\}$,
  $\mathcal{Q} \leftarrow \{\bm{p}_0\}$\;

  \While{$\mathcal{Q}$ is not empty}{
    $\bm{p} \leftarrow \text{\pop{$\mathcal{Q}$}}$\;

    \tcp{Extend the map forward.}
    $\mathcal{F} \leftarrow \text{\frontWaypoints{$\bm{p}$, $r_0$}}$\;
    $\bm{p}_f \leftarrow \text{\onRoute{$\bm{p}$, $\mathcal{F}$}}$\;

    \If{$\bm{p}_f$ exists}{
      \If{$\bm{p}_f \notin \mathcal{V}$ and $\text{\range{$\bm{p}$}}+ r_0 \leq r_m$}{
        $\text{\range{$\bm{p}_f$}} \leftarrow \text{\range{$\bm{p}$}}+ r_0$\;
        $\mathcal{V} \leftarrow \mathcal{V}\cup\{\bm{p}_f\}$,
        $\mathcal{Q} \leftarrow \mathcal{Q}\cup\{\bm{p}_f\}$\;
      }
      \lIf{$\bm{p}_f \in \mathcal{V}$}{
        $\mathcal{E} \leftarrow \mathcal{E}\cup\{(\bm{p}, \bm{p}_f)\}$
      }
    }



    \tcc{Extend the map to the left and right with similar steps using interfaces leftWaypoint and rightWaypoint.}
    $\vdots$\;
  }

  \caption{Directed-graph Map Construction}
  \label{alg: directed graph map construction}
\end{algorithm}

\begin{figure}
  \centering
  \subfloat[]{\includegraphics[width=0.3\columnwidth]{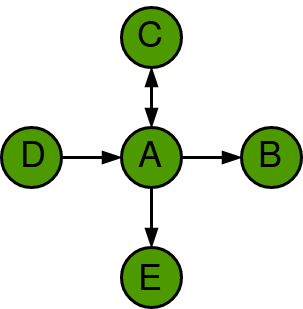}\label{subfig: microscopic directed graph map}}\hspace{2em}
  \subfloat[]{\includegraphics[width=0.3\columnwidth]{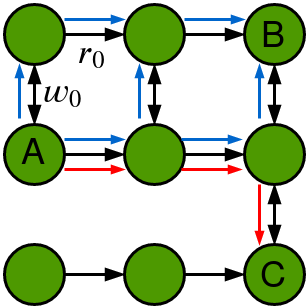}\label{subfig: illustration for functions d phi}}
  \caption{(a) shows a mini directed-graph map example demonstrating the connections of a vertex, $A$, with its four neighbor vertices. $A$, $B$, and $D$ are on the same lane. $D \rightarrow A \rightarrow B$ is the direction of the traffic. Therefore, the connections between $A$, $B$, and $D$ are one-way. $C$ and $E$ are on adjacent lanes of $A$. A two-way connection between $A$ and $C$ implies that vehicles can change lanes both from $A$ to $C$ and $C$ to $A$. In contrast, lane changing is only allowed from $A$ to $E$. (b) is an illustration for the function $d(\cdot)$ and $\phi(\cdot)$. In (b), $w_0$ and $r_0$ are the metric length of the lateral and longitudinal edges. The length of shortest path from both $A$ to $B$ and $A$ to $C$, is $w_0+2r_0$, \textit{i.e.} $d(A, B)=d(A, C)=w_0+2r_0$. However, $\phi(A, B)=3$, \textit{i.e} there are three paths (shown in \textcolor{blue}{blue}) from $A$ to $B$ with the shortest length, while $\phi(A, C)=1$ (shown in \textcolor{red}{red}).}
  \label{fig: microscopic directed graph map}
\end{figure}

\begin{figure}
  \centering
  \includegraphics[width=\columnwidth]{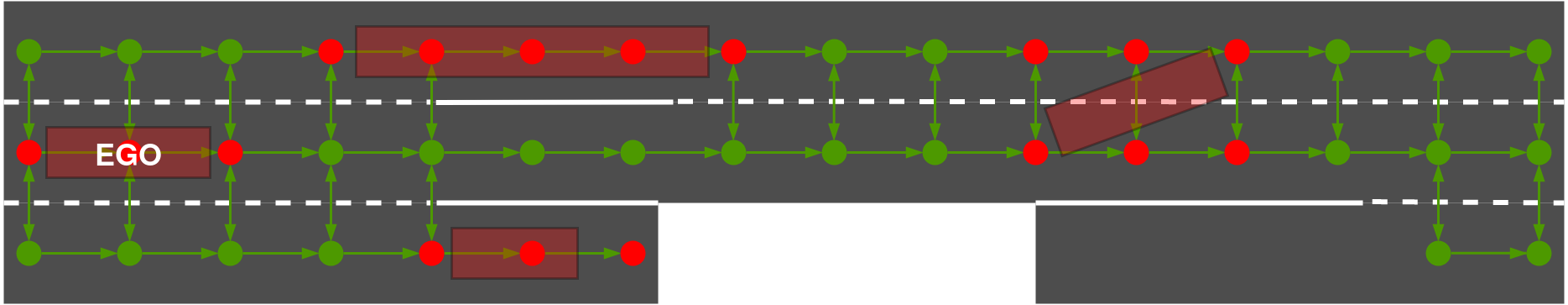}
  \caption{A directed-graph map overlaid on a highway environment. The right lane (lower) is discontinuous because of the off-ramp and on-ramp. In the middle section of the roads, left (upper) and center (middle) lanes are separated with solid lines preventing lane changes. Therefore, waypoints on adjacent lanes are not connected. No vertex is constructed at the initial section of the on-ramp on the right lane since this section is inaccessible to a vehicle already on the highway. Vehicles, shown as transparent \textcolor{red}{red} rectangles, are registered at the corresponding vertices, shown as \textcolor{red}{red} dots. If a vehicle is in the process of changing lanes, it occupies vertices on both lanes.}
  \label{fig: directed graph map}
\end{figure}

The proposed directed-graph map, $\mathcal{G} = \prl{\mathcal{V}, \mathcal{E}}$, consists of vertices, $\mathcal{V}$, and directed edges, $\mathcal{E}$. Vertices are waypoints, consisting of position, orientation, and curvature, regularly sampled along the lane centers. Directed edges model the connectivity between vertices. Each vertex in $\mathcal{V}$ can at most connect to its four neighbors. A direct edge from vertex $A$ to $B$ means that a vehicle can ``hop'' from a waypoint at $A$ to $B$ without violating any traffic rules, \textit{i.e.} only kinematics need to be considered. \figurename~\ref{subfig: microscopic directed graph map} shows a mini directed-graph map example demonstrating possible connections between vertices. \figurename~\ref{fig: directed graph map} shows an example of the directed-graph map on a highway environment of a larger scale.

The map construction requires interfacing with the map file (e.g. with OpenDrive format) of the environment at a larger scale and route information. Only four queries need to be made repeatedly: given a waypoint $\bm{p}$ at a lane center, 1) what are the accessible waypoints if a vehicle moves forward $x$ meters from $\bm{p}$; 2) what is the waypoint to the left of $\bm{p}$; 3) what is the waypoint to the right of $\bm{p}$; 4) is $\bm{p}$ on the pre-defined route. The first three queries are related to the environment and the last is to the route. Both are assumed to be known a priori. Alg.~\ref{alg: directed graph map construction} shows the construction of the directed-graph map.

The directed-graph map can be updated incrementally, extended, shortened, or shifted (a combination of extension and shortening). This can be achieved by maintaining the set of entrance and exit vertices. Entrance vertices are the ones with no vertex connecting to it from the back. Exit vertices are the ones with no vertex connected to its front. The procedure of extending the map is similar to Alg.~\ref{alg: directed graph map construction}. The difference is that $\mathcal{Q}$, in Alg.~\ref{alg: directed graph map construction}, is initialized with the set of exit vertices, instead of just the starting waypoint $\bm{p}_0$. Shortening the map is slightly different. Vertices, with the associated edges, have to be removed starting from the set of entrance vertices. The removal process stops until the desired range is met.

The directed-graph map can also be used to register the position and orientation of vehicles. Based on the state and the length of a vehicle, corresponding vertices in the directed graph are occupied. \figurename~\ref{fig: directed graph map} shows a directed-graph map with registered vehicles. Using the registered map, relative positions of the vehicles can be easily extracted. For example, one can follow the vertices on the same lane as the ego to identify its leading/following vehicle or its left and right leaders/followers using the vertices on adjacent lanes.

In order to specify the constraints (2) and (3) in Eq.~\eqref{eq: general autonomous driving problem}, we define a few functions operating on the directed-graph map. For all $\bm{p}\in\mathcal{V}$, define $s(\bm{p})$ and $l(\bm{p})$ as the coordinates of $\bm{p}$ in the Frenet frame of the road curve. In the case that $\bm{p}\notin\mathcal{V}$, define $s(\bm{p})=s(\bm{q})$ and $l(\bm{p})=l(\bm{q})$ where $\bm{q}$ is the projection of $\bm{p}$ on the graph, \textit{i.e.} $\bm{q}\in\mathcal{V}$ and $\|\bm{q}-\bm{p}\|_2 \leq \|\bm{q}'-\bm{p}\|_2$ for all $\bm{q}'\in\mathcal{V}$. Define $d(\bm{p}, \bm{q})$ as the length of the shortest path connecting $\bm{p}$ and $\bm{q}$, where path, in this case, is a sequence of edges in $\mathcal{E}$. Since there might be more than one shortest path, $\phi(\bm{p}, \bm{q})$ is defined as the number of shortest paths between $\bm{p}$ and $\bm{q}$. If either $\bm{p}$ or $\bm{q}$ is not in $\mathcal{V}$, $d$ and $\phi$ take their projections on the graph. \figurename~\ref{subfig: illustration for functions d phi} provides examples to illustrate the definitions of $d(\cdot)$ and $\phi(\cdot)$. Finally, we define the function $\mathcal{O}\prl{\bm{x}(t)}$, mapping the state of a vehicle to the set of vertices occupied by the vehicle. Note that the vehicle length is also required to determine the occupied vertices, but not included in $\bm{x}(t)$. However, we keep the notation $\mathcal{O}\prl{\bm{x}(t)}$ for brevity.

%% file: tex/algorithm.tex
\section{Algorithm}
\label{sec: algorithm}

\begin{algorithm}[t]
  \DontPrintSemicolon
  \SetCommentSty{emph}
  \SetKwProg{Fn}{Function}{}{end}
  \KwIn{
    $\bm{x}_s(t_0)$: the initial traffic state.\\
    \hspace{12mm}$\mathcal{G}$: the directed-graph map.}
  \KwOut{
    $\bm{p}_k$'s: the optimal motion primitives.}

  \SetKwFunction{pop}{pop}
  \SetKwFunction{frontEndPoint}{frontEndPoint}
  \SetKwFunction{leftFrontEndPoint}{leftEndPoint}
  \SetKwFunction{rightFrontEndPoint}{rightEndPoint}
  \SetKwFunction{satisfyConstraints}{satisfyConstraints}
  \SetKwFunction{extendGraph}{extendGraph}
  \SetKwFunction{terminals}{terminals}
  \SetKwFunction{optimalTerminal}{optimalTerminal}
  \SetKwFunction{backtrace}{backtrace}
  \SetKwFunction{dynamicalPath}{dynamicalPath}
  \SetKwFunction{simulate}{simulate}

  $\mathcal{S} \leftarrow \{(\bm{x}_s(t_0), \bm{p}_0, 0)\}$,
  $\mathcal{Q} \leftarrow \{(\bm{x}_s(t_0), \bm{p}_0, 0)\}$\;

  \While{$\mathcal{Q}$ is not empty}{
    $\bm{x}_s, \bm{p}, c \leftarrow \text{\pop{$\mathcal{Q}$}}$\;
    \tcp{Lane keep option.}
    $\bm{p}' \leftarrow \text{\frontEndPoint{$\bm{p}$, $\mathcal{G}$}}$\;
    \If{\satisfyConstraints{$\bm{p}, \bm{p}'$, $\mathcal{G}$}}{
      $\mathcal{S}, \mathcal{Q} \leftarrow \text{\extendGraph{$\bm{x}_s, \bm{p}, c, \bm{p}', \mathcal{S}, \mathcal{Q}$}}$
    }
   \tcp{Examine left and right lane change options.}
   $\vdots$\;
  }

  $\mathcal{T} \leftarrow \text{\terminals{$\mathcal{S}$}}$\;
  $\bm{x}_s, \bm{p}, c \leftarrow \text{\optimalTerminal{$\mathcal{T}$}}$\;
  \KwRet{\backtrace{$\bm{x}_s, \bm{p}, c$}}\;\;

  \Fn{$\mathcal{S}, \mathcal{Q} \leftarrow \text{\extendGraph{$\bm{x}_s, \bm{p}, c, \bm{p}', \mathcal{S}, \mathcal{Q}$}}$}{
    $\bm{\sigma} \leftarrow \text{\dynamicalPath{$\bm{p}, \bm{p}'$}}$\;
    $\bm{x}''_s, \bm{p}'', c_g \leftarrow \text{\simulate{$\bm{x}_s, \bm{\sigma}$}}$\;
    \lIf{collision}{\KwRet{$\mathcal{S}, \mathcal{Q}$}}
    \If{$\bm{p}''$ is $\bm{p}'$}{
      $\mathcal{Q} \leftarrow \mathcal{Q} \cup \{(\bm{x}''_s, \bm{p}'', c+c_g)\}$\;
    }
    $\mathcal{S} \leftarrow \mathcal{S} \cup \{(\bm{x}''_s, \bm{p}'', c+c_g)\}$\;
    \KwRet{$\mathcal{S}, \mathcal{Q}$}\;
  }

  \caption{Searching for Optimal Motion Primitives}
  \label{alg: graph search for optimal motion primitives}
\end{algorithm}

In this section, we introduce the solution to the problem in Sec.~\ref{sec: problem formulation}, and compare the runtime complexity of FELP against~\cite{ziegler2009spatiotemporal, mcnaughton2011motion, ajanovic2018search}. Since the complexity of FELP grows exponentially with the spatial horizon, we propose two variants with polynomial runtime complexity.

Alg.~\ref{alg: graph search for optimal motion primitives} shows the solution algorithm to the problem in Sec.~\ref{sec: problem formulation}. Starting from each waypoint $\bm{p}$ in $\mathcal{Q}$, three possible options are evaluated including lane keep, left and right lane change. For each option, the evaluation starts by locating the end point $\bm{p}'$ of the path segment. A dynamically feasible path segment, $\bm{\sigma}$, is then constructed connecting $\bm{p}$ and $\bm{p}'$ using the optimization algorithm in~\cite{kelly2003reactive}. Once the path segment is available, traffic dynamics can be forward simulated from $\bm{x}_s$ at $\bm{p}$ to $\bm{x}''_s$ at $\bm{p}''$. Note that it is not necessary that $\bm{p}''$ overlaps with $\bm{p}'$, since the ego may not reach $\bm{p}'$ within finite duration (\textit{e.g.} the traffic congestion). If so, $\bm{x}''_s$, together with $\bm{p}''$, will not be added to $\mathcal{Q}$ for further extension, but is treated as a terminal state. Once the spatial horizon is reached, the optimal terminal traffic state is identified, backtracing from which produces the optimal sequence of motion primitives. An example of the constructed paths is shown in \figurename~\ref{subfig: the optimal solution}. 

\begin{figure}[t]
  \centering
  \subfloat[Alg.~\ref{alg: graph search for optimal motion primitives} on~\cite{mcnaughton2011motion}]{\includegraphics[width=0.5\columnwidth]{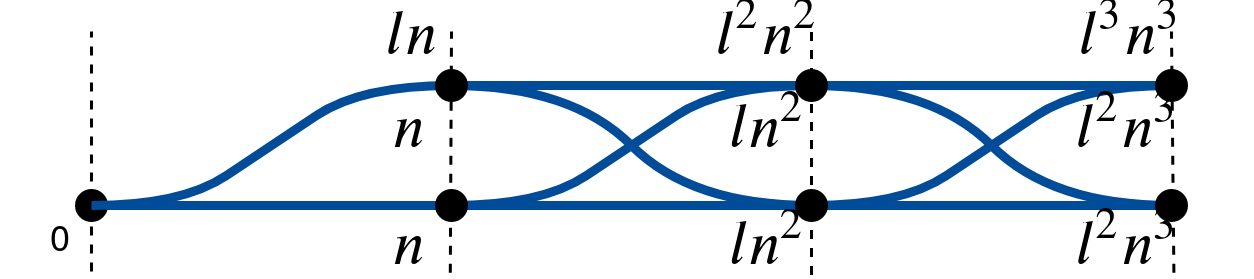}
              \label{subfig: spatiotemporal lattice complexity}}
  \subfloat[FELP]{\includegraphics[width=0.5\columnwidth]{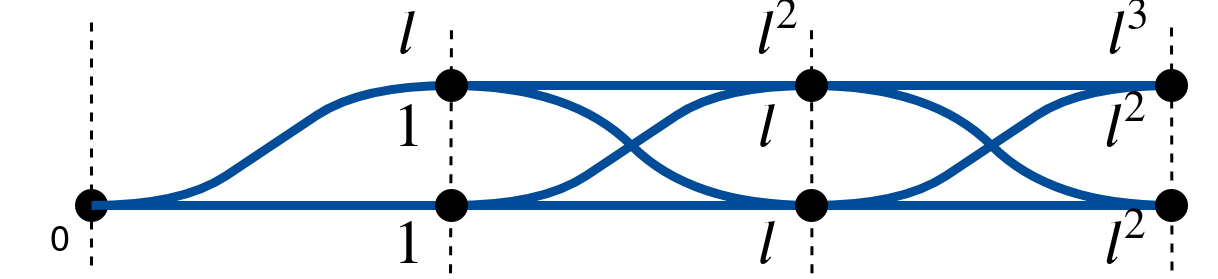}
              \label{subfig: speed feedback enhanced spatial lattice complexity}}\\
  \subfloat[C-FELP]{\includegraphics[width=0.5\columnwidth]{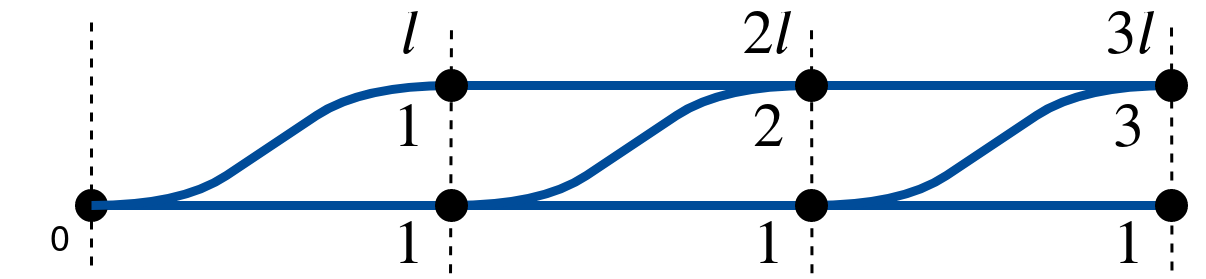}
              \label{subfig: constrained spatial lattice complexity}}
  \subfloat[R-FELP]{\includegraphics[width=0.5\columnwidth]{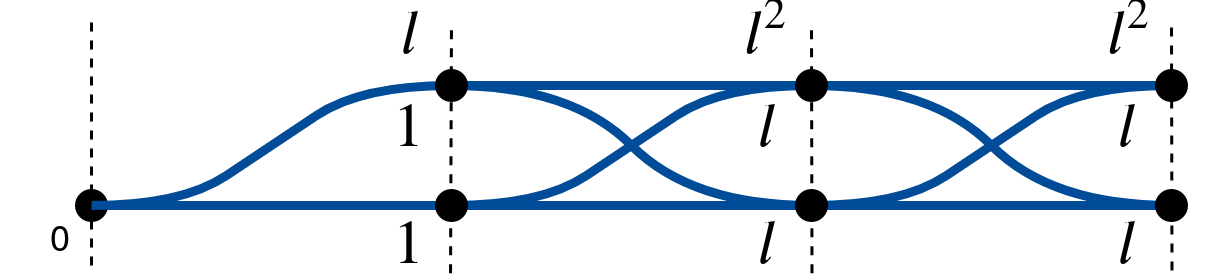}
              \label{subfig: pruned spatial lattice complexity}}
  \caption{(a), (b), (c), and (d) shows evaluated trajectories for Alg.~\ref{alg: graph search for optimal motion primitives} applied to the problem in~\cite{mcnaughton2011motion}, Sec.~\ref{sec: problem formulation}, and its two variants in Sec.~\ref{subsec: additional constraints} and Sec.~\ref{subsec: removing traffic states} respectively. Each figure shows the paths (\textcolor{blue}{blue}) created for a road with $l$ lanes, although only two lanes are drawn. The vertical dash lines marks the discretization of the longitudinal dimension. Next to each node (\textbf{black}) marks the number of trajectory segments ending at the node, the sum of which for the nodes on the same column is marked at the top. The total number of evaluated trajectories can be obtained by summing the values on the top row of each figure.}
  \label{fig: lattice planner complexity}
\end{figure}

In the following, we compare the runtime complexity of FELP with~\cite{mcnaughton2011motion}, \cite{ziegler2009spatiotemporal}, and \cite{ajanovic2018search}. The runtime complexity is quantified with the number of evaluated trajectories. Variables that determine the complexity include the discretization density of the 2-D position, 2-D velocity, 2-D acceleration, and time. $n$ and $l$ denote the density of the discretization along the longitudinal and lateral directions of a road, where $l$ is often less than $n$. To avoid cluttering of expressions, we use $n$ to also denote the discretization density of other dimension.

In~\cite{ziegler2009spatiotemporal}, there are $ln^6$ nodes in the lattice (In order to avoid confusion with the vertices in the directed-graph map in Sec.~\ref{subsec: directed graph map representation}, we use ``node'' to refer to the vertices in the lattice). Assuming the trajectories going out from a node only connect to the nodes one unit away in the longitudinal dimension, the out-degree of a node is $ln^5$. Therefore, the total number of trajectories to be evaluated is $ln^6 \cdot ln^5 = O\prl{l^2 n^{11}}$.

Both~\cite{mcnaughton2011motion} and~\cite{ajanovic2018search} use hybrid A*~\cite{dolgov2008practical} to reduce runtime complexity, which prunes the induced position, velocity, or time using a pre-defined grid. Hybrid A*, as an approximation to graph search algorithms, may fail to find the optimal solution. Therefore, we consider the complexity of using authentic graph search algorithms, such as Alg.~\ref{alg: graph search for optimal motion primitives}, for the problems in~\cite{mcnaughton2011motion} and~\cite{ajanovic2018search}. As discussed in Sec.~\ref{sec: introduction}, both~\cite{mcnaughton2011motion} and~\cite{ajanovic2018search} share the same runtime complexity. We focus only on~\cite{mcnaughton2011motion} because of its similarity with this work. \figurename~\ref{subfig: spatiotemporal lattice complexity} shows the growth of the number of evaluated trajectories as the spatial horizon extends. The total number of trajectories is $\sum_{k=1}^n l^k n^k = O(l^n n^n)$. Note that since $n$ is often less than $10$, the running time of~\cite{mcnaughton2011motion} may not be more than~\cite{ziegler2009spatiotemporal}.

\figurename~\ref{subfig: speed feedback enhanced spatial lattice complexity} shows the growth of the trajectories of FELP. The total number of trajectories is $\sum_{k+1}^n l^k = O(l^n)$. The significant reduction of the runtime complexity is due to the usage of IDM as the speed feedback policy in the ego motion primitives. Unlike~\cite{mcnaughton2011motion}, one no longer has to discretize acceleration, evaluate different constant accelerations over the same path, producing more trajectories. Instead, in FELP, the trajectory of the ego is uniquely determined by its initial state and a path. Constructing paths requires a spatial lattice only. Although the runtime complexity is significantly reduced compared to~\cite{mcnaughton2011motion}, it still grows exponentially with spatial horizon, $n$. In the following, we propose two methods to further reduce the runtime to polynomial.

\input{./tex/algorithm/additional_constraints}
\input{./tex/algorithm/removing_traffic_states}

%% file: tex/algorithm/additional_constraints.tex
\subsection{Introducing Additional Constraints (C-FELP)}
\label{subsec: additional constraints}
One possible way to reduce the runtime complexity of a graph search is introducing additional constraints, thereby eliminating a significant number of options. In C-FELP, we introduce a new restriction for the ego where at most one lane change is allowed over the entire planning horizon. Considering the spatial planning horizon for a local trajectory planner is on the order of one hundred meter, the new restriction is sensible. Effectively, this is equivalent to replacing the constraint $|l(\bm{p}_k) - l(\bm{p}_{k-1})| \leq w_0$  in Eq.~\eqref{eq: unconstrained autonomous driving problem} with $|l(\bm{p}_k) - l(\bm{p}_{0})| \leq w_0$.
All path end points are either on the same lane or adjacent lanes to $\bm{p}_0$. \figurename~\ref{subfig: constrained spatial lattice complexity} shows the growth of trajectories with the new constraint. The total number of constructed trajectories is $\sum_{k=1}^n kl = O(n^2l)$.

%% file: tex/algorithm/removing_traffic_states.tex
\subsection{Removing Traffic States (R-FELP)}
\label{subsec: removing traffic states}
\begin{algorithm}[t]
  \DontPrintSemicolon
  \SetCommentSty{emph}
  \SetKwProg{Fn}{Function}{}{end}
  \SetKwFunction{snapshotAt}{snapshotAt}
  \SetKwFunction{removeSnapshotAt}{removeSnapshotAt}

  \Fn{$\mathcal{S}, \mathcal{Q} \leftarrow \text{\extendGraph{$\bm{x}_s, \bm{p}, c, \bm{p}', \mathcal{S}, \mathcal{Q}$}}$}{
    $\bm{\sigma} \leftarrow \text{\dynamicalPath{$\bm{p}, \bm{p}'$}}$\;
    $\bm{x}''_s, \bm{p}'', c_g \leftarrow \text{\simulate{$\bm{x}_s, \bm{\sigma}$}}$\;
    \lIf{collision}{\KwRet{$\mathcal{S}, \mathcal{Q}$}}

    \If{$\bm{p}''$ is not $\bm{p}'$}{
      $\mathcal{S} \leftarrow \mathcal{S} \cup \{(\bm{x}''_s, \bm{p}'', c+c_g)\}$\;
      \KwRet{$\mathcal{S}, \mathcal{Q}$}\;
    }

    \If{$\bm{p}'' \in \mathcal{S}$}{
      $\bm{x}'''_s, \bm{p}'', c''' \leftarrow \text{\snapshotAt{$\mathcal{S}, \bm{p}''$}}$\;
      \lIf{$c_g+c \geq c'''$}{\KwRet{$\mathcal{S}, \mathcal{Q}$}}
      $\mathcal{S} \leftarrow \text{\removeSnapshotAt{$\mathcal{S}, \bm{p}''$}}$\;
      $\mathcal{Q} \leftarrow \text{\removeSnapshotAt{$\mathcal{Q}, \bm{p}''$}}$\;
    }

    $\mathcal{S} \leftarrow \mathcal{S} \cup \{(\bm{x}''_s, \bm{p}'', c+c_g)\}$\;
    $\mathcal{Q} \leftarrow \mathcal{Q} \cup \{(\bm{x}''_s, \bm{p}'', c+c_g)\}$\;
  }

  \caption{Modified extendGraph}
  \label{alg: extending the graph with pruning}
\end{algorithm}

The underlying reason of the $O(l^n)$ runtime complexity of FELP is the fast growing of traffic states at the nodes. Different traffic states at a node are created if the ego can reach the node through different combination of paths. A direct way to address this is to ensure that only one traffic state is maintained at each node. We propose to use the cost-to-come of trajectories to determine which traffic state should be maintained at the nodes. Alg.~\ref{alg: extending the graph with pruning} is an update of the function \verb!extendGraph! reflecting this idea. \figurename~\ref{subfig: pruned spatial lattice complexity} shows that the runtime complexity of the modified algorithm is $\sum_{k=1}^n l^2 = O(nl^2)$. It should be noted that, although the complexity is reduced compared to $O(l^n)$, the principle of optimality is violated. To see this, the optimal sequence of motion primitives may have higher cost-to-come compared to another sequence that reaches the same intermediate waypoint. As a result, the optimal sequence is removed at an early stage of the algorithm, and thus the returned sequence of motion primitives is non-optimal, in general.

%% file: tex/experiments.tex
\section{Experiment}
\label{sec: experiment}
In the experiments, We compare FELP and its variants with the spatiotemporal lattice planner in~\cite{mcnaughton2011motion} through a merging scenario and highway traffic simulations. In addition, we show the performance of FELP with different traffic densities. For ease of reference, we call the method in~\cite{mcnaughton2011motion} STLP, which is the most similar to FELP in terms of the planning methodology. In our implementation of STLP, six accelerations are provided, $\{-8, -4, -2, -1, 0, 1\}m/s^2$. To ensure reasonable runtime, hybrid-A* is applied. Longitudinal speed is discretized to three intervals, while the temporal dimension is left undiscretized.
All simulations are built with CARLA~\cite{Dosovitskiy17}. The planning algorithms are implemented to run with a single CPU core. The reported timing of the following experiments are obtained with Intel Core i9-9920X running at 3.5GHz. Implementations of the algorithms in this work can be found at \url{https://github.com/KumarRobotics/conformal\_lattice\_planner}.
\input{./tex/experiments/merging_scenario}
\input{./tex/experiments/highway_traffic}
\input{./tex/experiments/traffic_density}

%% file: tex/experiments/merging_scenario.tex
\subsection{Merging Scenario}
\label{subsec: merging scenario}
\begin{figure}[t]
  \centering
  \includegraphics[width=\columnwidth]{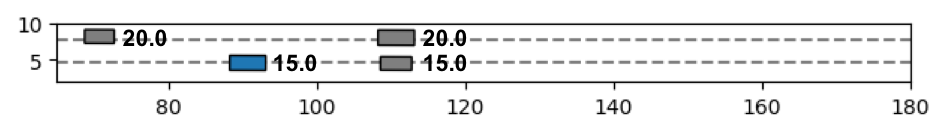}  \\
  \includegraphics[width=\columnwidth]{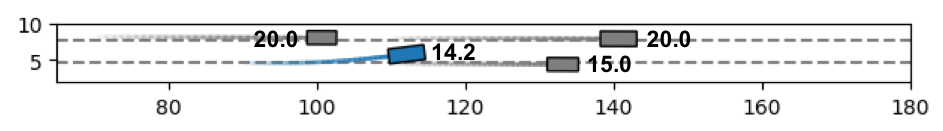} \\
  \includegraphics[width=\columnwidth]{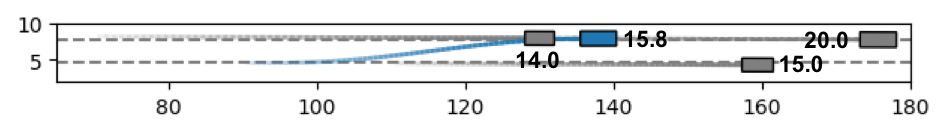} \\
  \caption{The merging scenario where the ego (\textcolor{blue}{blue}) merges into the left lane. The three snapshots shows the traffic at different time, $0.0s$, $1.5s$, $3.25s$ respectively. Next to each vehicle marks the speed ($m/s$) of the vehicles at the corresponding time.}
  \label{fig: merging scenario}
\end{figure}
In this scenario, the ego is moving at $15m/s$, $20m$ behind a leading vehicle with the same speed, trying to merge to the left. Two vehicles on the left lane are faster moving at $20m/s$, $20m$ ahead and behind the ego respectively. The scenario simulates a typical on-ramp merging case. Since agent vehicles are assumed to move with constant velocity in STLP, lane changing will cause a collision in prediction, and thus is not allowed at this moment. As a result, the ego may fail to merge and has to take the exit. However, lane changing is possible with FELP as agents are modeled with IDM. \figurename~\ref{fig: merging scenario} shows traffic snapshots with the ego planned with FELP.

As discussed in Sec.~\ref{sec: problem formulation}, the assumption that agents would always yield may lead to inconsiderate behavior of the ego. Therefore, the cost induced by agents' braking is used to prevent aggressive actions from the ego. In this example, one can tune the cost function in case more conservative behavior is desired. However, tuning the cost function does not help in producing a lane-change option in STLP.

%% file: tex/experiments/highway_traffic.tex
\subsection{Highway Traffic}
\label{subsec: highway traffic}
\begin{table*}
  \centering
  \begin{threeparttable}
    \renewcommand{\arraystretch}{1.5}
    \caption{Comparison of different methods with one-hour highway traffic simulations}
    \label{tab: highway traffic simulation results}
    \begin{tabular}{c | c | c | c | c | c | c }
      \hline\hline
      & jerk\tnote{1} ($m/s^3$) & acceleration\tnote{1} ($m/s^2$) & speed\tnote{1} ($m/s$) & headway\tnote{1, 2}\hspace{1em} ($s$) & induced brake\tnote{3} ($m/s^2$) & planning time ($ms$)\\ \hline
      STLP & $-10.00/ 10.00$ & $-4.00/ 1.00$ & $16.70/ 20.00$ & $1.15/ 4.70$ & $2.00$ & $1631$ \\
      \hline
      FELP & $-0.43/ 0.51$ & $-0.52/ 0.54$ & $15.45/ 20.00$ & $1.23/ 4.75$ & $3.04$ & $276$ \\
      \hline
      C-FELP & $-0.36/ 0.36$ & $-0.41/ 0.38$ & $16.04/ 20.00$ & $1.29/ 4.80$ & $3.04$ & $153$ \\
      \hline
      R-FELP & $-0.35/ 0.38$ & $-0.41/ 0.34$ & $15.95/ 20.00$ & $1.26/ 4.75$ & $1.79$ & $232$ \\
      \hline\hline
    \end{tabular}
    \begin{tablenotes}
      \item [1] The data represents $1\%$ and $99\%$ percentile.
      \item [2] Headway is computed whenever there is a leading vehicle for the ego. Otherwise, headway is not defined.
      \item [3] Induced brake refers to brake of the follower on the target lane when the ego changes lanes. The data represent the $1\%$ percentile.
    \end{tablenotes}
  \end{threeparttable}
\end{table*}

The highway traffic simulation is set on the map Town04, a default map in CARLA. To simulate the limited perception range of the ego, traffic is maintained in the neighborhood of the ego vehicle, specifically, $100m$ ahead and $50m$ behind the ego. Within the $150m$ range, $8$ agent vehicles are maintained. In the case that an agent vehicle moves out of the range, a new vehicle is added close to either the end or the front of the perception range, simulating the detection of new vehicles. As in Sec.~\ref{sec: problem formulation}, all agent vehicles are lane followers with the speed modulated by IDM. The desired speed of the agent vehicles are set to around $20m/s$, close to the desired speed of the ego. In order to improve the similarity with real-world scenarios, behavior of agents are randomized through two ways. First, the IDM hyper-parameters of the agents deviate from nominal values. Second, the online desired speed of agent vehicles is perturbed with slow varying Gaussian noise.

Table~\ref{tab: highway traffic simulation results} reports the performance of different methods with a one-hour highway traffic simulation. Average values for the metrics are almost the same for all methods, thus omitted. Instead, the percentile data are shown to reflect the extreme statistics. A few observations can be made. Comparing jerk and acceleration, FELP and its variants are less aggressive, implying an improvement in the comfort. This is because IDM produces continuous acceleration if no abrupt change is observed from the traffic. The proposed methods also significantly improve the running time agreeing with the analysis in Sec.~\ref{sec: algorithm}. According to the induced brake, FELP and its variants may expect harder braking from the followers on the target lanes when the ego changes lanes. As discussed in Sec.~\ref{subsec: merging scenario}, the induced brake can be reduced by tuning the objective function. Comparing C-FELP, R-FELP, and FELP, the performance is almost the same per the metrics, although C-FELP and R-FELP generates sub-optimal trajectories in terms of the objective function. However, C-FELP and R-FELP do further reduce the planning time, which makes them more suitable for online usage.

%% file: tex/experiments/traffic_density.tex
\subsection{Effect of the Traffic Density}
\label{sec: different traffic densities}
\begin{table*}
  \centering
  \begin{threeparttable}
    \renewcommand{\arraystretch}{1.5}
    \caption{Performance of FELP in one-hour highway traffic simulations with different traffic density\tnote{1}}
    \label{tab: felp with different traffic density results}
    \begin{tabular}{c | c | c | c | c | c | c }
      \hline\hline
      agent \# & jerk ($m/s^3$) & acceleration ($m/s^2$) & speed ($m/s$) & headway ($s$) & induced brake ($m/s^2$) & planning time ($ms$)\\ \hline
      $4$ & $-2.87\times10^{-3}/ 0.0$ & $0.00/ 8.28\times10^{-3}$ & $19.92/ 20.00$ & $1.80/ 4.80$ & $1.35$ & $239$ \\
      \hline
      $8$ & $-0.43/ 0.51$ & $-0.52/ 0.54$ & $15.45/ 20.00$ & $1.23/ 4.75$ & $3.04$ & $276$ \\
      \hline
      $12$ & $-0.50/ 0.61$ & $-0.60/ 0.54$ & $15.18/ 20.00$ & $1.21/ 4.78$ & $2.62$ & $317$ \\
      \hline\hline
    \end{tabular}
    \begin{tablenotes}
      \item [1] Data is in the same format as Table~\ref{tab: highway traffic simulation results}.
    \end{tablenotes}
  \end{threeparttable}
\end{table*}
The performance of FELP is also compared with different traffic densities. The setup of the experiments is similar to that in Sec.~\ref{subsec: highway traffic}. The difference is the number of agent vehicles is configured to $4$, $8$, and $12$ respectively to simulate various traffic densities. Table~\ref{tab: felp with different traffic density results} reports the experimental results. As the number of agent vehicles increases, both the comfort (reflected by jerk and acceleration) and the safety (reflected by headway and induced brake) of the trajectories are affected, which agrees with the normal driving experience. It is also worth noting that the planning time of FELP increases linearly with the traffic density, making FELP suitable for applications in dense traffic scenarios.

%% file: tex/conclusion.tex
\section{Conclusion}
\label{sec: conclusion}
In this work, we propose a new lattice planning approach, FELP. FELP applies IDM as the speed feedback policy to modulate the speed of the ego and predict the behavior of agents. IDM enables the responsive behavior of agent vehicles. We show that such modeling can prevent over-conservative behavior of the ego in a merging scenario simulation. Combining IDM with paths, we are able to construct ego motion primitives as feedback policies. With the velocity and acceleration determined by the motion primitives, only spatial dimensions need to be discretized. The reduction in the lattice dimensionality significantly improves the efficiency of FELP. We show this by comparing runtime complexity and actual online planning time of FELP with other lattice planners.

Two directions of extending the current work are promising. First, FELP remains a deterministic planning approach, relying on receding horizon control to cope with the mismatch between the prediction and actual scenarios. Stochasticity can be introduced in modeling the motion of agent vehicles. Planning with the new model promises the capability of considering various possible future scenarios. Another future direction is extending the current work to urban environments. Compared with highways, urban driving is more challenging because of more complicated traffic rules and more types of dynamical obstacles in addition to agent vehicles.